\documentclass[review]{elsarticle}

\usepackage{lineno,hyperref}
\modulolinenumbers[5]
\journal{arxiv}

\begin{document}

\begin{frontmatter}

\title{SACDNet: Towards Early Type 2 Diabetes Prediction with Uncertainty for Electronic Health Records}

\author[1]{Tayyab Nasir}
\cortext[mycorrespondingauthor]{Corresponding author}
\ead{tayyabnasir22@gmail.com}

\author[2]{Muhammad Kamran Malik}
\address[1]{CureMD Research and Development}
\address[2]{Department of Information Technology, Faculty of Computing and Information Technology, University of the Punjab, Lahore, PK}

\begin{abstract}
Type 2 diabetes mellitus (T2DM) is one of the most common diseases and a leading cause of death. The problem of early diagnosis of T2DM is challenging and necessary to prevent serious complications. This study proposes a novel neural network architecture for early T2DM prediction using multi-headed self-attention and dense layers to extract features from historic diagnoses, patient vitals, and demographics. The proposed technique is called the Self-Attention for Comorbid Disease Net (SACDNet), achieving an accuracy of 89.3\% and an F1-Score of 89.1\%, having a 1.6\% increased accuracy and 1.3\% increased f1-score compared to the baseline techniques. Monte Carlo (MC) Dropout is applied to the SACDNet to get a bayesian approximation. A T2DM prediction framework based on the MC Dropout SACDNet is proposed to quantize the uncertainty associated with the predictions. A T2DM prediction dataset is also built as part of this study which is based on real-world routine Electronic Health Record (EHR) data comprising 4,124 diabetic and 181,767 non-diabetic examples, collected from 295 different EHR systems running in different parts of the United States of America.  This dataset is further used to evaluate 7 different machine learning and 3 deep learning-based models. Finally, a detailed analysis of the fairness of every technique against different patient demographic groups is performed to validate the unbiased generalization of the techniques and the diversity of the data.

\end{abstract}

\begin{keyword}
\texttt Artificial neural networks\sep Deep Learning\sep Corpus generation\sep Healthcare\sep Uncertainty\sep Disease Prediction\sep 

\end{keyword}

\end{frontmatter}

\section{Introduction}
\label{sec:introduction}
Diabetes mellitus or simply diabetes is a set of conditions concerning the body’s mechanism of managing blood glucose~\cite{typediabetes, sun2022idf}. There are two major types of diabetes mellitus i.e., Type 1 Diabetes mellitus (T1DM) and Type 2 Diabetes mellitus (T2DM)~\cite{typediabetes, sun2022idf}. Almost 90\% of the diabetics are type 2 diabetics which signifies the importance of diagnosing T2DM~\cite{typediabetes, sun2022idf}. T2DM occurs more often in middle-aged and older people~\cite{typediabetes, sun2022idf, abhari2019artificial}. However, over the past 20 years, this disease is becoming more common in adolescents and even in children~\cite{sun2022idf}. 

Symptoms of T2DM include unplanned weight loss, increased thirst (polydipsia), frequent urination (polyuria), and increased hunger~\cite{wei2018comprehensive, sun2022idf, sonar2019diabetes, ramachandran2014know}. Other symptoms of T2DM involve slow wound healing, fatigue, depression, and high blood pressure~\cite{howlader2022machine, sun2022idf}. Previous studies show higher comorbidity of T2DM with various diseases including cardiovascular disease, hypertension, angina, gastric ulcer, and hypothyroidism~\cite{xiong2019machine, hossain2020framework, kim2012comorbidity, hassing2004comorbid, ducat2014mental}. It can cause serious long-term complications like affecting the kidneys, eyes, and nerves~\cite{haq2020intelligent, alhassan2018type}. Also, it increases the risk of heart disease and stroke~\cite{diabetes2015long}. Similarly, there are several risk factors involved when it comes to T2DM including obesity, ethnicity, older age, high blood pressure, family history, and genetics~\cite{fazakis2021machine, collins2011developing, robertson2011blood}.

According to the International Diabetes Federation (IDF), around 536.6 million people were diabetic in the year 2021~\cite{sun2022idf}. This number is projected to grow to 783.2 million by 2045. As stated almost 90\% of total diabetics are diagnosed with T2DM. Thus, the total number of type 2 diabetic persons in the year 2021 was 483 million~\cite{sun2022idf}. Diabetes has a high mortality rate as well. According to the IDF, 6.7 million people died due to some complications caused by diabetes in 2021 alone~\cite{sun2022idf}. The USA is also impacted hugely by diabetes. A total of 32.2 million US citizens were diabetic in the year 2021, which amounts to almost 11\% of the total US population~\cite{sun2022idf, nationaldiabetesreport}. The US is at number 4 on the list of the countries with the most diabetic patients and spends a fortune every year fighting diabetes and its underlying complications~\cite {sun2022idf}. In 2021 a total of 966 billion US dollars were spent on fighting diabetes which is 316\% higher than the 232 billion US dollars spent in the year 2007 for the same~\cite{sun2022idf}.

Early detection of diabetes is of vital importance for an effective treatment~\cite{franciosi2005use} yet stays a challenge~\cite{zhu2020deep}. T2DM is often diagnosed very late when the symptoms become severe and complications arise~\cite {sun2022idf, franciosi2005use}. Diagnosis of diabetes is usually delayed even in developed countries. Early diagnosis of T2DM can help prevent many serious complications including but not limited to loss of vision, kidney failure, amputations of limbs, stroke, and even premature death~\cite{krasteva2011oral, jayanthi2017survey}. Thus, an early diagnosis of T2DM is essential which can not only save millions of dollars every year but also save precious human lives. However, diabetes in general and T2DM, in particular, remain undiagnosed in the vast majority~\cite{franciosi2005use}, and according to the National Diabetes Statistics Report, around 8.5 million diabetic persons in the US alone are undiagnosed~\cite{nationaldiabetesreport}.

Predicting T2DM has been studied previously using supervised and unsupervised machine learning algorithms~\cite{taz2021comparative, mani2012type, zou2018predicting, ducat2014mental, alhassan2018type}. Most of the research work has used PIMA Indian diabetes dataset for training and evaluating T2DM predictive systems~\cite{ghosh2021comparative, joshi2021predicting, birjais2019prediction, yahyaoui2019decision, rahman2020deep}. PIMA dataset has a very limited number of examples, which raises the concern of a lack of versatility in the dataset that is required for better generalization of the models. Also, the dataset contains examples belonging to only the female PIMA Indians of Arizona further limiting the variation in the dataset. Also, some other datasets like the UCI and MIMIC-III have been used but such datasets are collected in a more controlled environment and may lack the inherent features and characteristics of real-world data.

This work aims to overcome the shortcomings of the existing research including the lack of volume and diversity of datasets, data being collected in a controlled environment, and the nature of features like tricep skin thickness and blood insulin that are not part of routine EHR data. To this extent following key contributions are made as part of this study:
\begin{itemize}
\item A novel deep neural network architecture is proposed for the diagnosis of T2DM. The proposed technique called the SACDNet uses multi-headed self-attention to extract relational features from historic diagnoses and uses fully connected dense layers to extract features from vitals and demographics. The technique then uses a set of fully connected layers to predict T2DM using these features.
\item A novel framework is proposed to predict T2DM with associated confidence that can help in identifying uncertain predictions. The proposed framework is built using the proposed SACDNet with MC-Dropout and entropy to get uncertainty associated with the predictions.
\item A novel dataset is built collecting data from different EHR systems running in different parts of the US. The dataset encloses a set of diverse data points that can be used to train a more generalized T2DM predictive model.
\item Different machine and deep learning techniques were used to train and evaluate T2DM predictive models using the proposed corpus, building T2DM predictive systems that rely solely on routine EHR data features.
\item Also, a fairness analysis of all the techniques used in the study is presented, evaluating each against different demographic groups of data, verifying the claimed diversity of the proposed dataset and the fairness of each technique.
\end{itemize}

The rest of the paper is divided as follows: Section~\ref{sec:relatedwork} presents related work. Section~\ref{sec:datasetgeneration} gives an overview of the corpus generation process and its specifications. Section~\ref{sec:methodology} presents the novel deep neural network-based technique used in this study, along with the details of the proposed framework for diabetes prediction. The details of the experimentation are given in section~\ref{sec:experiments}. The results obtained and the analysis performed on the results are presented in Section~\ref{sec:resultsandanalysis}. Section~\ref{sec:conclusionfuturework} provides the conclusions and directions for future work.

\section{Related Work}
\label{sec:relatedwork}
Different machine and deep learning techniques have been previously used for the diagnosis of type 2 diabetes~\cite{yahyaoui2019decision, rahman2020deep, mani2012type, zou2018predicting, ducat2014mental, alhassan2018type}. Many datasets have been used for training and evaluating machine learning techniques for T2DM prediction. Three significant datasets have been used previously for T2DM detection which are the PIMA Indian Diabetes dataset~\cite{smith1988using}, UCI~\cite{strack2014impact}, and the MIMIC-III~\cite{johnson2016mimic}. Out of these, the PIMA Indian diabetes dataset is the most widely used. Table~\ref{tab:table1} gives the details of each dataset. 

\begin{table*}
\centering
\small
\caption{Details of existing datasets.}
\label{tab:table1}
\begin{tabular}{ l l l l}
\hline
Dataset &	Positive Examples &	Negative Examples &	Total Examples \\ \hline\hline 
PIMA &	179 &	358 &	537 \\ \hline
UCI &	51,034 &	14,806 &	65,840 \\ \hline
MIMIC-III &	1,242 &	38,456 &	39,698 \\ \hline\hline

\end{tabular}
\end{table*}

A large number of studies have used the PIMA dataset to build T2DM prediction models~\cite{alhassan2018type}. The very first study that used machine learning for predicting T2DM using the PIMA dataset dates back to 1988~\cite{smith1988using}. Techniques like KNN, Logistic Regress, Naïve Bayes, Decision Trees, Support Vector Machines (SVM), Random Forests, AdaBoosting, and Gradient Boosted Trees have all been used to train and evaluate T2DM prediction models using the PIMA dataset~\cite{taz2021comparative, ghosh2021comparative, kalagotla2021novel, battineni2019comparative, mujumdar2019diabetes, hasan2020diabetes, kahramanli2008design, kaur2014improved, maniruzzaman2018accurate, tigga2020prediction, sarwar2018prediction, prema2019prediction, birjais2019prediction, joshi2021predicting}. Apart from the conventional machine learning techniques, deep neural networks have also been widely used for T2DM prediction. Previous studies have used multi-layer perceptron networks to predict T2DM, where PIMA was used for training and evaluation~\cite{temurtas2009comparative, wei2018comprehensive, ayon2019diabetes, naz2020deep, nadesh2020type}. Many other neural network architectures including CNN, RNN (both LSTM and GRU), and a combination of CNN-LSTM-based deep neural networks were evaluated using the PIMA dataset~\cite{yahyaoui2019decision, rahman2020deep, garcia2021diabetes, alhassan2018type}. Additionally, many studies have used voting and stacking-based ensembling to combine the aforementioned techniques to improve the overall performance of T2DM prediction models~\cite{taz2021comparative, kayaer2003medical, kalagotla2021novel, prema2019prediction}.

Previous studies have also applied machine and deep learning techniques to datasets other than PIMA ~\cite{lee2015identification, haq2020intelligent, fazakis2021machine, xiong2019machine, tigga2020prediction, sonar2019diabetes, el_jerjawi2018diabetes, ducat2014mental}. For example, NHANES dataset~\cite{ncfh2017national} was used to train an SVM-based model for predicting diabetes~\cite{yu2010application}. Similarly, another study combined NHANES dataset~\cite{ncfh2017national} with lab results of different patients for training Logistic Regression, SVM, Random Forest, and Gradient Boosted Trees for predicting T2DM~\cite{kayaer2003medical}. Many studies have also collected data consisting of the same features as the PIMA dataset from other sources to build machine learning-based T2DM predictive models~\cite{haq2020intelligent, mujumdar2019diabetes}. 

Over the past decade, a large number of hospitals and practices have moved to digital health records~\cite{johnson2016mimic, cheng2016risk}. Previous work has used data from different EHRs to build machine learning models~\cite{robertson2011blood, mani2012type, zheng2017machine, kopitar2020early} for T2DM prediction. For example, studies have used Logistic Regression, SVM, Random Forest, Naïve Bayes, and Gradient Boosted Trees on EHR data collected from different hospitals in different parts of the world~\cite{dagliati2018machine, farran2013predictive, viloria2020diabetes}. Also, deep neural networks have been trained on EHR data for T2DM prediction~\cite{krasteva2011oral, alhassan2018type}. Most such studies that utilize some EHR data rely on clinical components like lab results, historical diagnoses, family medical history, or a combination of these. Although data collected from EHRs might be significantly larger yet in most of the studies data is collected from a single practice or region and might not be an effective sample of the data required for better generalization of the machine and deep learning techniques~\cite{lai2019predictive, deberneh2021prediction, cahn2020prediction}.

In addition to the aforementioned techniques that rely solely on tabular data features, some of the studies also rely on images and signal data for the prediction of T2DM. A couple of studies have utilized heart rate variation data collected from ECG and built machine and deep learning models to predict diabetes~\cite{swapna2018automated, swapna2018diabetes}. Similarly, images of the retina have been used by a couple of studies to predict T2DM using deep CNN and LSTM-based models~\cite{islam2021dianet, samant2018machine}.

\section{Dataset}
\label{sec:datasetgeneration}
For this study, the data was provided by CureMD\footnote{https://www.curemd.com/} which is one of the leading EHR providers in the US. Data from 767 different practices was provided, out of which 295 unique practices were selected based on the existence of type 2 diabetic patients. Hence it can be stated that the dataset that is used in this study has examples belonging to 295 different EHR environments, operating in various parts of the US. The motivation behind using such a dataset is to be able to build real-world T2DM prediction systems that rely on real-world features coming from routine EHR records, which cannot only provide a non-invasive method of early T2DM detection but can also be plugged into any EHR, without requiring any additional data from patients.

The raw data from CureMD’s EHRs contained 5,092,987 unique clinical encounters belonging to 1,034,889 different patients. All the data attributes that contained personal information related to the patients were already masked and are not used in any of the experiments or analyses. For this study only the following three clinical components are used:

\begin{itemize}
  \item Diagnosis: This clinical component consists of a set of ICD-10-CM codes that are assigned to a patient during a clinical encounter. Each code represents a disease, symptom, or injury\footnote{https://www.cdc.gov/nchs/icd/index.htm} that was found in the patient at the time of the encounter. The raw data from the 295 different practices contained 29,714 unique ICD-10-CM codes. This clinical component is selected because it provides the key information related to the individual disease that can either lead to or is comorbid with T2DM.
  \item Vitals: The vitals clinical component represents a set of vitals readings that were taken during a visit. The aforementioned EHR data contained 17 different vitals including weight, height, BMI, lean body weight, ideal body weight, neck circumference, waist, oxygen saturation, peak expiratory flow, blood type, blood Rh, finger stick, pulse, respiration, temperature, systolic blood pressure, and diastolic blood pressure. This clinical component was selected because many previous studies relied on the use of vitals for predicting the chances of developing T2DM. Vitals like weight, BMI, blood pressure, etc., have all been marked as important markers in T2DM prediction~\cite{rahman2020deep, el_jerjawi2018diabetes, krasteva2011oral, kayaer2003medical, deberneh2021prediction, yu2010application}.
  \item Demographics: The patient’s demographics component consists of attributes including the patient’s date of birth (which can be used to calculate the present age of the patient), gender, sexual orientation, and race category. Demographical information has been widely used in various disease prediction systems and is considered an important feature~\cite{yu2010application, fazakis2021machine, lu2022patient, joshi2018diabetes}.
\end{itemize}

The CureMD EHR data in its raw form required a certain set of preprocessing steps to mold it into a form usable for experiments which are discussed in the section below.

\subsection{Data Preprocessing}
\label{subsec:datapreprocessing}
Data collected from EHRs is sparse, noisy, heterogeneous, and unstructured~\cite{zheng2017machine, rashidian2020detecting, cheng2016risk}. Such data also presents other challenges including the problem of missing data and class imbalance~\cite{shivade2014review, hripcsak2013next}. The raw EHRs data that was obtained from CureMD had all of these inherent issues. The details of these data challenges and the required preprocessing techniques that were used to tackle these challenges are discussed below:

\subsubsection{Handling Records with no History}
The aim of this study is to build an early T2DM prediction system that relies solely on patients’ historic clinical diagnoses, recorded vitals, and demographics. Thus, it is desired to filter out only those patients that have some historic visits before being diagnosed with T2DM. However, in the raw EHR data, a large number of records were encountered where a patient was marked with T2DM (indicated with an E11 ICD-10-CM code) in the very first encounter. Such examples occur in cases when a patient, already diagnosed with T2DM walks into a hospital or practice for the first time without any recorded medical history in the EHR of the same. Thus, a strategy is required to eliminate such example points having no historic encounters prior to the diagnosis of T2DM. For this, all the T2DM-diagnosed patients that do not have at least 4 previous encounters before being diagnosed with T2DM were filtered out of the data. A total of 72,626 patients with T2DM were identified in the raw data, out of which only 5,770 had at least previous 4 encounters before being diagnosed with T2DM. For the non-diabetics, the patients with less than 4 encounters were filtered out. 

After the execution of this step, the data set was reduced from 72,626 to 5,770 T2DM patients. Also, from a total of 962,263 non-diabetic patients, only 186,865 patients were left. Hence, after executing the first step of our preprocessing a dataset with 5,770 diabetic and 186,865 non-diabetic patients was created.

\subsubsection{Handling Missing Data}
The problem of missing data is widely encountered in EHRs~\cite{shivade2014review, hripcsak2013next}. Different studies have implied different approaches for handling missing data. Techniques like dropping complete records with missing values, and filling in missing values using mean, median, or from the nearest neighbors, have been used in many research studies~\cite{ghosh2021comparative, mujumdar2019diabetes, xiong2019machine, maniruzzaman2018accurate, deberneh2021prediction, birjais2019prediction, zou2018predicting}. Raw EHR records were analyzed to identify the attributes that were most affected by the problem of missing values. As a result of this analysis, it was found that all the attributes that have the problem of missing values belonged to the vitals component. To deal with missing values a 2-step strategy was applied. In the first step, a missing value ratio was calculated for every attribute using equation \ref{eq:eq1}.

\begin{equation}
\label{eq:eq1}
Missing Values Ratio = \frac{Number Of Records With Missing Values}{Total Number of Records} * 100
\end{equation}

\begin{table*}
\centering
\small
\caption{Missing value ratio for vitals.}
\label{tab:table2}
\begin{tabular}{ l l}
\hline
Attribute &	Missing Value Ratio (\%) \\ \hline\hline 
Weight &	12.67 \\ \hline
Height &	21.53 \\ \hline
BMI &	28.34 \\ \hline
Lean Body Weight &	64.71 \\ \hline
Ideal Body Weight &	65.14 \\ \hline
Neck Circumference &	92.22 \\ \hline
Waist &	91.80 \\ \hline
Oxygen Saturation &	89.15 \\ \hline
Peak Expiratory Flow &	99.99 \\ \hline
Blood Type &	99.82 \\ \hline
Blood Rh &	99.94 \\ \hline
Finger Stick &	98.98 \\ \hline
Pulse &	28.91 \\ \hline
Respiration &	53.93 \\ \hline
Temperature &	51.66 \\ \hline
Systolic Blood Pressure &	16.48 \\ \hline
Diastolic Blood Pressure &	16.44 \\ \hline\hline
\end{tabular}
\end{table*}

Table~\ref{tab:table2} presents the details of the missing value ratio of different attributes in the vitals component. Based on the results presented in table~\ref{tab:table2} the vitals attributes for which the missing value ratio was higher than 50\% were dropped. The reason for dropping these attributes with a higher missing value ratio lies in the fact that such variables do not possess a sufficient amount of information that can be used to impute the missing values in the next step. 

The second step of the applied strategy consists of imputing the missing values for the remaining vitals attributes. The following set of steps was used to impute the missing values for vitals:
\begin{itemize}
  \item Firstly, all the records were grouped based on the patient they belong to.
  \item Secondly, the patient groups having no value for any of the vitals attributes were dropped, as the missing data cannot be imputed for such patients.
  \item Thirdly, for each patient group, the Exponentially Weighted Moving Average (EWMA) of every vitals attribute was calculated and stored.
  \item Finally, all the missing values were populated by their respective calculated values, calculated using the EWMA.
\end{itemize}
EWMA was used for data imputation keeping in mind the fact that EHR data is temporal in nature. EWMA can be used to assign larger weights to recent observations, hence giving more importance to the recent ones as compared to the older data points~\cite{holt2004forecasting}. Such behavior is desirable and quite effective in capturing the temporal nature associated with a patient’s EHR records where more weight is given to recent visits and less weight to older visits. For the remaining two components i.e., patient demographics and diagnosis the problem of missing value did not exist in the raw data. After applying this 2-step strategy the dataset size was further reduced to a total of 4,124 diabetic and 181,767 non-diabetic patients.

\subsubsection{Creating Single Example Points}
The data of every patient consists of multiple encounters/visits. This step deals with merging these multiple encounters of a single patient into one example point that can be used for training and evaluating machine and deep learning techniques. The process of creating the example points is given as follows:
\begin{itemize}
  \item All three components i.e., diagnosis, vitals, and demographics, were merged into a single record for every patient encounter.
  \item For every encounter the age of the patient at the time of the encounter was calculated using the patient’s date of birth.
  \item Next for every patient all the encounter records were sorted using the date of the encounter in ascending order of the date.
  \item For every T2DM patient, all the encounter records before the first T2DM diagnosis encounter were selected and the associated diagnoses were extracted. The ICD-10-CM diagnosis codes were merged together creating a single comma-separated string enclosing all the unique historic diagnoses of a patient.
  \item Next, the vitals from the first T2DM encounter were merged with the diagnoses attribute along with the demographic information of the patient, creating a single T2DM positive example. 
  \item Label 1 was also generated for the created T2DM example points.
  \item For every non-T2DM patient, all the diagnoses from all the previous encounters were merged to create a single comma-separated diagnoses feature. Also, the vitals from the last encounter and the patient’s demographic information were merged with the diagnoses feature to create a T2DM negative example.
  \item Label 0 was generated for the non-T2DM example points.
\end{itemize}

\subsubsection{Dealing with Class Imbalance}
Class imbalance is a problem that arises when the data is distributed unevenly among the classification categories~\cite{vluymans2019learning, chawla2002smote, japkowicz2000class}. In case of class imbalance, the performance of traditional classification techniques is affected where the performance of the techniques becomes biased towards the majority class and is reduced for the minority class~\cite{vluymans2019learning, lakshmi2014study, bunkhumpornpat2011mute}. Many techniques have been presented over the years to deal with class imbalance including oversampling, undersampling, SMOTE, MUTE, ADASYN~\cite{bunkhumpornpat2011mute, chawla2002smote, ling1998data, kubat1997addressing, he2008adasyn}. Class imbalance is another inherent property of health care and medicine data~\cite{wang2019dmp_mi, deberneh2021prediction, krasteva2011oral}.

Techniques like SMOTE, ADASYNC, data oversampling, data undersampling, etc. have been used for balancing data for T2DM prediction as well~\cite{krasteva2011oral, garcia2021diabetes, deberneh2021prediction}. The EHR data that is used as part of this study has a huge class imbalance having a total of 4,124 diabetic and 181,767 non-diabetic records. A combination of 5-fold cross-validation combined with undersampling is applied to handle this class imbalance. A total of 5 non-overlapping samples of size 4,124 were selected from the non-T2DM examples which are used to train 5 different variations of every machine learning and deep learning technique. The details are discussed in section~\ref{sec:experiments}.

\section{Methodology}
\label{sec:methodology}
A novel deep neural network architecture is proposed for T2DM prediction that combines multi-headed self-attention with fully connected layers. The proposed technique called the Self-Attention for Comorbid Disease Net (SACDNet) uses multiheaded self-attention to extract features from the diagnoses vector with the intent to find a relation between every pair of diagnoses and uses dense layers to extract features from vitals and demographics data. The resultant feature vectors are then concatenated and passed through a single dropout layer. Finally, a series of fully connected layers are used for T2DM classification.

\subsection{Self Attention}
\label{subsec:selfattention}
The concept of attention was introduced to modify the RNN-based Encoder-Decoder models~\cite{bahdanau2014neural} where every decoder timestamp will now receive a weighted sum of all encoder hidden states called the attention score. The attention score is usually a scaled dot product given by the equation \ref{eq:eq2}.

\begin{equation}
\label{eq:eq2}
Attention(Q,K,V)= Softmax(\frac{QK^{T}}{\sqrt{d_{m}}})V
\end{equation}
Where Q, K, and V represent the query, key, and value matrixes, obtained by multiplying the inputs sequence X with three different weight matrixes $W_{q}$, $W_{k}$, and $W_{v}$ respectively. If the query, key, and value are all generated using the same input sequence then the above equation \ref{eq:eq2} becomes self-attention. The concept of self-attention involves finding the similarity between all elements in the input sequence, in our case, this will be to find a relation between all historic diagnoses.

\subsection{Multi-Headed Self Attention}
\label{subsec:multiselfattention}
This involves applying self-attention to the input sequence multiple times, obtaining different weight matrixes for the query, key, and value~\cite{vaswani2017attention}. Thus, it provides a more diverse set of relationships between the elements of the input sequence. These multiple self-attention outputs are concatenated. For given input sequence X multi-headed self-attention for N heads is given by equation \ref{eq:eq3}.

\begin{equation}
\label{eq:eq3}
MHSA(Q,K,V)=Concat(Att_{1}(Q,K,V), Att_{2}(Q,K,V), ...., Att_{N}(Q,K,V)))W_{O}
\end{equation}
Where $Att_{i}$ represents self-attention obtained using different $W_{q}$, $W_{k}$, and $W_{v}$ weight metrixes. N represents the number of heads.

\subsection{SACDNet}
\label{subsec:sacdnet}
For the given diagnosis matrix D, and the vitals/demographics matrix V SACDNet feature extraction is given using equation \ref{eq:eq4}.

\begin{equation}
\label{eq:eq4}
Feature(D,V)=Concat(MHSA(Q,K,V),MLP(V))
\end{equation}

Where MLP represents a set of fully connected layers. This feature vector is then passed to another set of fully connected layers, for T2DM classification. The details of the network are given in table~\ref{tab:table3}.

\begin{table*}
\centering
\footnotesize
\caption{Architecture of SACDNet.}
\label{tab:table3}
\begin{tabular}{ l p{5.3cm} p{5.3cm}}
\hline
No. &	\multicolumn{2}{c}{\textbf{Network Layers}} \\ \hline\hline 
1 & Diagnosis Input Vector &	Vitals/Demographics Input Vector \\ \hline
2 & Multiheaded Self Attention (N=3, Query Dimension=3, Key Dimension=3, Value Dimension=3) & Fully Connected (Units=256, Activation=SELU) \\
3 & & Fully Connected (Units=128, Activation=SELU) \\ \hline
4 & \multicolumn{2}{c}{Concatenate (To a single merged feature vector)} \\ \hline
5 & \multicolumn{2}{c}{Dropout (Rate=0.1)} \\ \hline
6 & \multicolumn{2}{c}{Fully Connected (Units=64, Activation=SELU)} \\ \hline
7 & \multicolumn{2}{c}{Fully Connected (Units=64, Activation=Sigmoid)} \\ \hline
\hline
\end{tabular}
\end{table*}

\subsection{Monte Carlo (MC) Dropout}
\label{subsec:mcdropout}
Deep neural networks are deterministic in nature, which is not always desirable, especially in the case of high-risk systems such as clinical systems~\cite{shridhar2019comprehensive, seedat2019towards, gawlikowski2021survey}. It is desirable to have a system that can provide associated confidence with its prediction. The probability distribution returned by the Softmax function cannot be considered a model’s confidence as it has been proven that neural network-based models can make incorrect predictions with higher confidence~\cite{gal2016dropout, szegedy2013intriguing, nguyen2015deep, gawlikowski2021survey}. Epistemic or Model uncertainty estimation using MC dropout~\cite{gal2016dropout} has been used widely for deep neural network-based models\cite{zhong2020environmental, nair2020exploring, da2022improving, roy2019bayesian}. The idea of MC-Dropout is to get a prediction for given input X from T highly similar ensemble models and then obtain an average of these T different output probability distributions \ref{eq:eq5}. This is achieved by applying the concept of dropout regularization at inference time, which in fact yields T different distributions of the model’s parameters. Hence, a large difference in the prediction of these T different models can be an indicator of larger model uncertainty and such can help in obtaining more robust predictive confidence~\cite{gawlikowski2021survey, nair2020exploring}.

\begin{equation}
\label{eq:eq5}
p(y=1)=\frac{1}{T}\sum_{i}^{T}P(y=1|x, W_{i})
\end{equation}

Where for given T forward passes from the network we will have T different parameter distributions {$W_{1}$, $W_{2}$, $W_{3}$, …., $W_{T}$} as given in equation \ref{eq:eq5}. 

\subsection{Measuring Predictive Uncertainty}
\label{subsec:measuringuncertainty}
For the given output probability distributions from T different passes, different measures have been used for uncertainty quantization~\cite{zhong2020environmental, gal2017deep, nair2020exploring}. Entropy has been widely used as a measure of uncertainty quantization~\cite{zhong2020environmental, gal2017deep, lukovnikov2021detecting}, where entropy for the averaged MC-Dropout results is calculated using the equation \ref{eq:eq6}.

\begin{equation}
\label{eq:eq6}
H(y) = -\sum_{c}^{C}\frac{1}{T}\sum_{t}^{T}P(y=c_{c}|x,W_{t})*log(\sum_{t}^{T}P(y=c_{c}|x,W_{t}))
\end{equation}

\subsection{Proposed Framework}
\label{subsec:proposedframework}
The proposed framework combines the novel SACDNet with MC-Dropout to build a system that can predict T2DM with associated uncertainty. The idea is to get an ensemble of 100 different highly similar models, for which an average prediction probability for T2DM is calculated \ref{eq:eq7}. 

\begin{equation}
\label{eq:eq7}
MC-SACDNet=\frac{1}{T}\sum_{t}^{T=100}SACDNet_{t}
\end{equation}

\begin{figure}[ht]
    \centering
    \includegraphics[width=135mm]{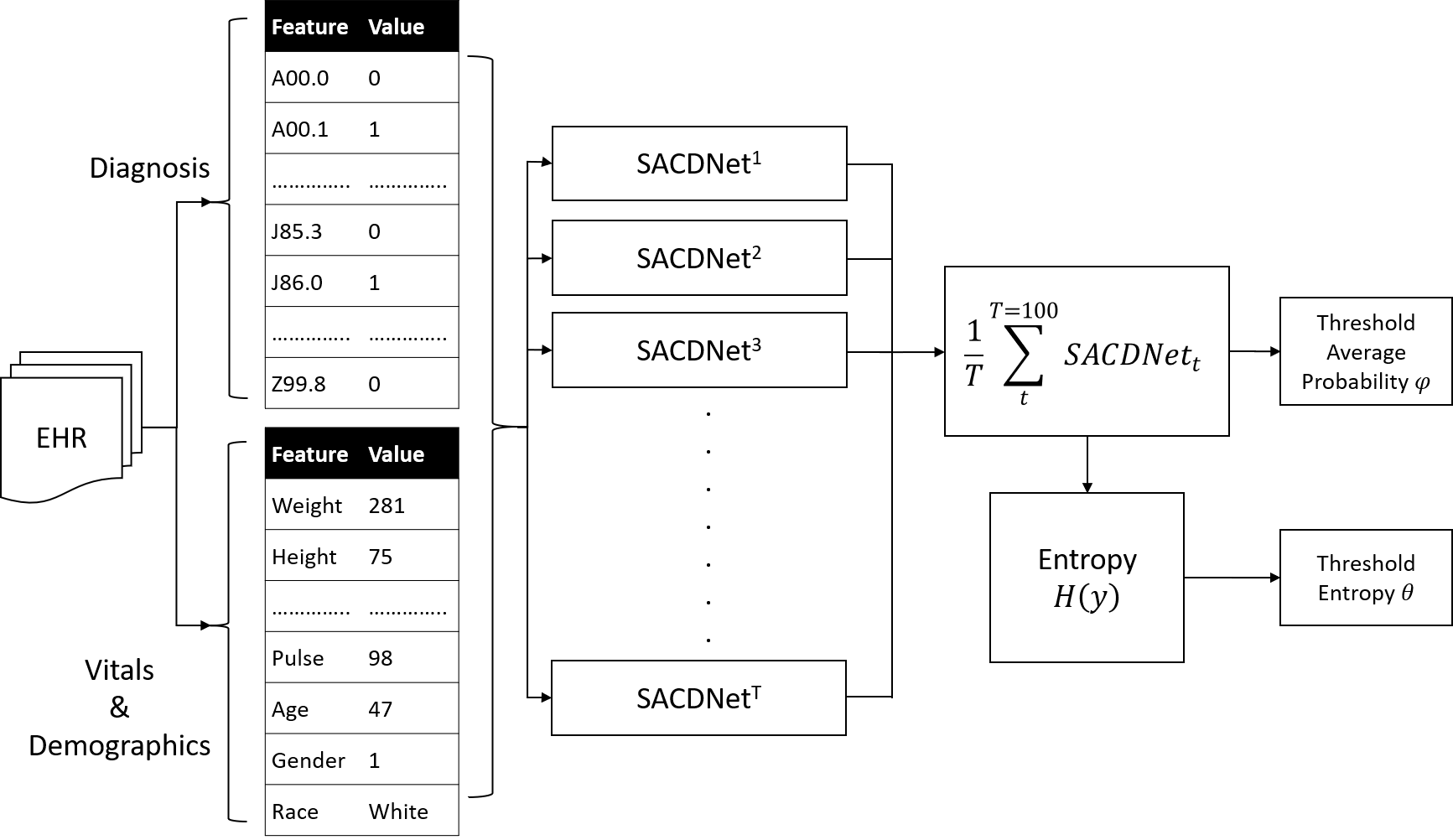}
    \caption{Proposed framework for T2DM prediction with uncertainty.}
    \label{fig:fig1}
\end{figure}

Next using the entropy metric discussed before uncertainty value for the prediction is calculated and based on a specified threshold theta a prediction is marked as certain or uncertain. An overview of the framework is given in figure~\ref{fig:fig1}. It can be observed in the figure~\ref{fig:fig1} the electronic health record of a patient is divided into 2 different feature vectors comprising of the historic diagnoses and vitals-demographics respectively. These feature vectors are passed through 100 different representations of the MC-Dropout SACDNet. The generated set of probabilities is averaged to get the final probability which is passed through a threshold function to predict T2DM. Also, the same mean probability is used to quantize uncertainty using entropy.

\section{Experiments}
\label{sec:experiments}
This section presents details of experiments performed on the developed T2DM corpus. Machine learning algorithms, deep learning algorithms, and proposed SACDNet are applied to the dataset to train T2DM classifiers. As discussed in section~\ref{sec:datasetgeneration}, the proposed corpus has a huge class imbalance. This class imbalance can affect the performance of the classifiers~\cite{vluymans2019learning, lakshmi2014study, bunkhumpornpat2011mute}. To deal with this problem 5-fold cross-validation was applied obtaining a balanced set by randomly undersampling the majority class. 7 different machine learning techniques including AdaBoost, Decision Trees, Logistic Regression, Gaussian Naïve Bayes, Random Forest, SVM, and Gradient Boosted Trees (XGB) were trained using the proposed corpus. Multiple deep neural network-based techniques were trained using the proposed corpus~\cite{naz2020deep, ayon2019diabetes, rahman2020deep, ayon2019diabetes, nadesh2020type}, however, apart from our proposed novel technique SACDNet, only two other techniques were able to produce satisfactory results, referred to as the FCN and FCN Dropout. The FCN network consists of 4 fully connected hidden layers using ReLU activations whereas the FCN Dropout consists of 3 fully connected hidden layers with hidden layer one having a dropout rate of 30\% and hidden layer two having a dropout rate of 20\%, all using the same ReLU activations.

The details of the experimentation process are given as follows:
\begin{itemize}
  \item Firstly, all the T2DM positive and negative class examples were separated.
  \item Next 5 random samples, without overlap were selected from the non-T2DM examples, where the size of each sample selected was equal to the size of our T2DM positive examples set.
  \item Next 5 different datasets were created combining the T2DM positive set with each of the 5 non-T2DM sets.
  \item Next for every dataset an 80-20\% train test split was done.
  \item Every technique was trained using a 5-fold cross-validation, where one of the 5 balanced samples was used for each fold.
  \item Performance of each model per fold was calculated using the evaluation measures discussed in the results section~\ref{sec:resultsandanalysis}.
  \item An average performance of every model was also calculated using the results from the 5 folds.
\end{itemize}
For a fair comparison, none of the models’ hyperparameters were tuned for both machine and deep learning techniques

\section{Results and Analysis}
\label{sec:resultsandanalysis}

\subsection{Evaluation Measures}
\label{subsec:evaluationmeasures}
For the evaluation of the models presented in this study, 5 widely used metrics were used including Accuracy, F1-Score, Precision, Recall, and Specificity. These have been used for evaluating disease prediction systems, especially for T2DM prediction~\cite{taz2021comparative, battineni2019comparative, sonar2019diabetes, hasan2020diabetes, kahramanli2008design, haq2020intelligent, howlader2022machine, maniruzzaman2018accurate, tigga2020prediction, birjais2019prediction, zou2018predicting, mani2012type}. For the evaluation of the uncertainty of the model, we have used entropy as discussed in the methodology section~\ref{sec:methodology}.

\subsection{Comparison Under Fixed Settings}
\label{subsec:comparisonfixed}
\begin{figure}[ht]
    \centering
    \includegraphics[width=125mm]{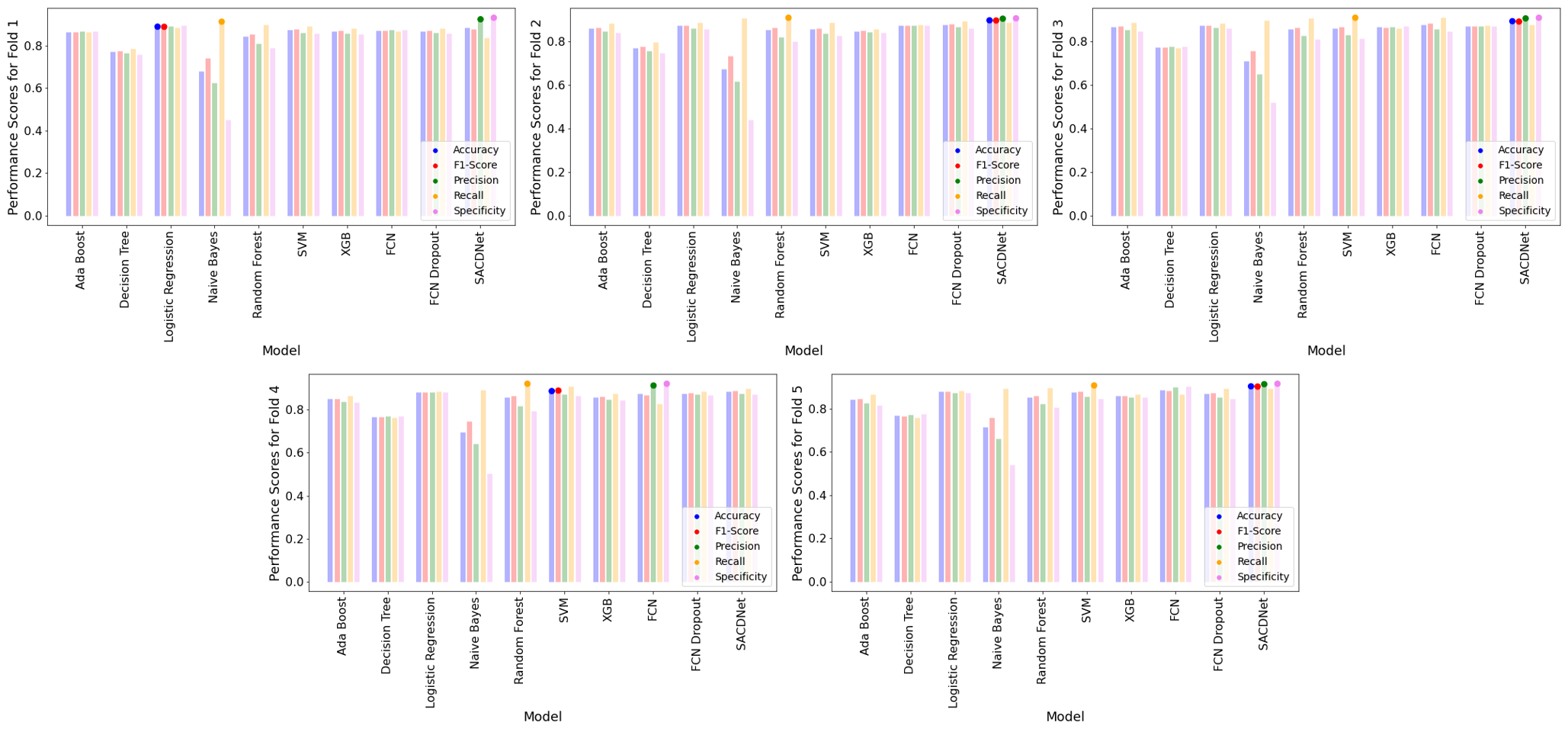}
    \caption{Results of 5 folds cross-validation.}
    \label{fig:fig2}
\end{figure}

This section presents a comparison of the performance of all the techniques discussed, using different evaluation measures. Firstly, the results obtained against the test set for each of the 5 data folds are presented in figure~\ref{fig:fig2}. Next, the average accuracy, f1-score, precision, recall, and specificity are presented in table~\ref{tab:table4}. 

\begin{table*}
\centering
\footnotesize
\caption{Mean of 5 folds cross-validation results.}
\label{tab:table4}
\begin{tabular}{ l l p{1.2cm} p{1.2cm} p{1.2cm} p{1.2cm} p{1.2cm}}
\hline
No. & Model &	Accuracy (\%) &	F1-Score (\%) &	Precision (\%) &	Recall (\%) &	Specificity (\%) \\ \hline\hline 
1 &	Ada Boost &	85.8 &	86.0 &	84.7 &	87.4 &	85.8 \\ \hline
2 &	Decision Tree &	77.1 &	77.2 &	76.8 &	77.5 &	77.1 \\ \hline
3 &	Logistic Regression &	87.9 &	88.0 &	87.5 &	88.5 &	87.9 \\ \hline
4 &	Naïve Bayes &	69.6 &	74.8 &	63.9 &	90.1 &	69.6 \\ \hline
5 &	Random Forest &	85.4 &	86.1 &	82.0 &	90.7 &	85.4 \\ \hline
6 &	SVM &	87.2 &	87.6 &	85.2 &	90.1 &	87.2 \\ \hline
7 &	XGB &	86.1 &	86.2 &	85.5 &	86.8 &	86.1 \\ \hline
8 &	FCN &	87.7 &	87.6 &	88.4 &	86.9 &	87.7 \\ \hline
9 &	FCN Dropout &	87.3 &	87.5 &	86.4 &	88.5 &	87.3 \\ \hline
10 & SACDNet &	89.3 &	89.1 &	90.5 &	87.9 &	89.3 \\ \hline\hline

\end{tabular}
\end{table*}

It can be observed from the table~\ref{tab:table4} that our proposed technique SACDNet performs better compared with 7 different machine learning and 2 neural network-based techniques, achieving a 1.6\% increased accuracy and 1.3\% increased F1-Score compared to the other techniques.

\subsection{Fairness Evaluation}
\label{subsec:fairnessevaluation}
It is desirable to have a T2DM prediction system that is able to perform equally well on a diverse set of examples, without having any bias towards a certain gender, age, or racial group. The proposed dataset collected from different EHR systems running in different parts of the US was built to capture examples belonging to a larger population. Such a diverse dataset can be helpful in building more generalized machine and deep learning models that can perform equally well for examples belonging to different demographic groups of patients. In order to evaluate this hypothesis, the next set of analyses was performed to validate the fairness of each of the techniques used. 

\begin{figure}[ht]
    \centering
    \includegraphics[width=100mm]{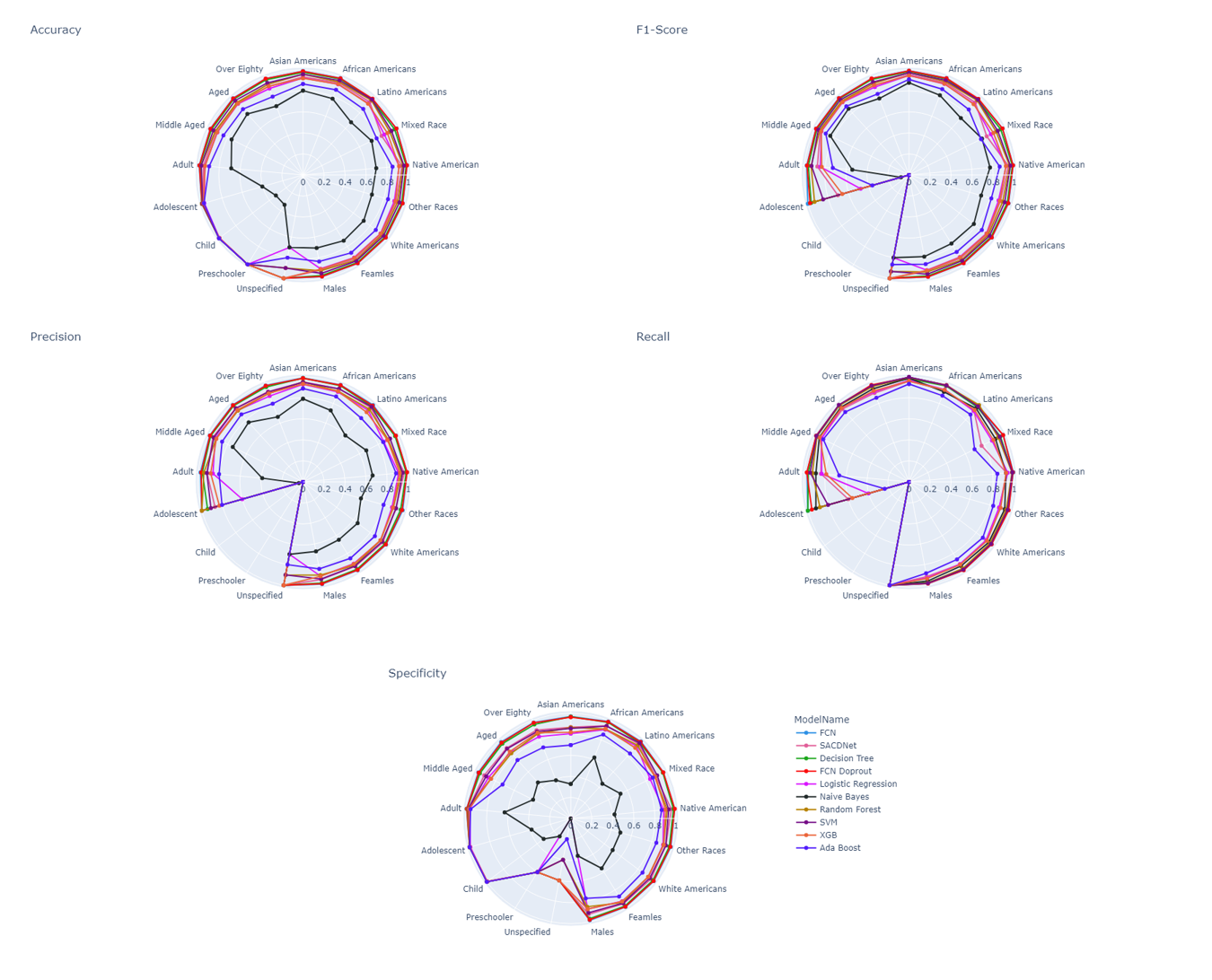}
    \caption{Performance of every technique against certain demographic groups.}
    \label{fig:fig3}
\end{figure}

For fairness evaluation, the proposed dataset was divided into certain groups based on the associated demographic information of the patients. Example points were divided based on age, gender, and race. All the techniques were evaluated against the individual groups and the results are presented in the figure~\ref{fig:fig3}. 

It can be observed from the figure~\ref{fig:fig3} that all the techniques have identical performance for all the race and gender groups. However, there is a sudden drop in the case of precision, recall, specificity, and f1-score for two age groups namely the children and preschoolers. The reason for this is the unavailability of significant examples belonging to the two aforementioned groups. For example, for preschoolers, there were only 1569 examples in the dataset all of which belonged to the non-diabetic class. Hence, there were no diabetic patients belonging to this class, and the precision, recall, specificity, and f1-score, cannot be calculated. Similarly, for Children, only 1857 examples were present all of which belonged to the non-diabetic class, and hence the same holds. Hence, this sudden drop does not represent the biasness of the techniques but only the unavailability of data for the calculation of precision, recall, specificity, and f1-score, which can be further validated by no drop in the accuracy for these two groups. Also, for the unspecified gender, only one example of a diabetic patient and only 6 examples of non-diabetics were encountered which again is causing discrepancies in the performance of certain evaluation measures even though the accuracy of the models is close to near perfect. Hence it can be stated that all the techniques were able to generalize fairly against all the data groups which confirm the diversity of the proposed dataset.

\subsection{Uncertainty Analysis}
\label{subsec:uncertaintyanalysis}
\begin{figure}[ht]
    \centering
    \includegraphics[width=55mm]{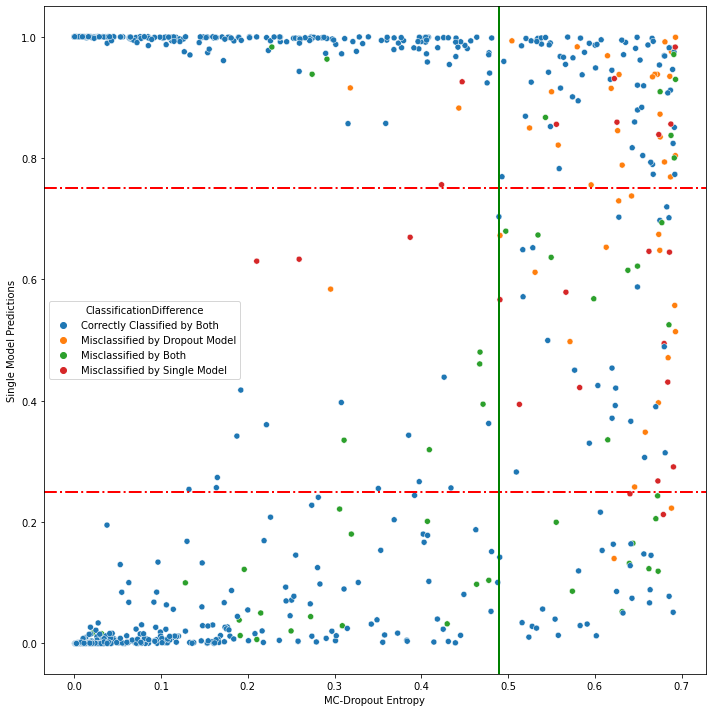}
    \caption{Comparison of confidences of SACDNet and MC-SACDNet for misclassified examples.}
    \label{fig:fig4}
\end{figure}

As discussed in the methodology section~\ref{sec:methodology}, the proposed SACDNet is intended to be part of a complete T2DM prediction framework that can make predictions with associated confidence, marking predictions with lower confidence as uncertain. In order to draw a comparison of the proposed technique and its MC-Dropout-based extension figure~\ref{fig:fig4} was plotted. The horizontal straight dashed lines represent the probability thresholding for the simple model while the vertical straight line represents the entropy threshold for the MC-Dropout extension. It can be observed from the figure~\ref{fig:fig4} that the MC-Dropout extension used with entropy is able to identify more misclassified examples as uncertain, making it more reliable. 

\begin{figure}[ht]
    \centering
    \includegraphics[width=85mm]{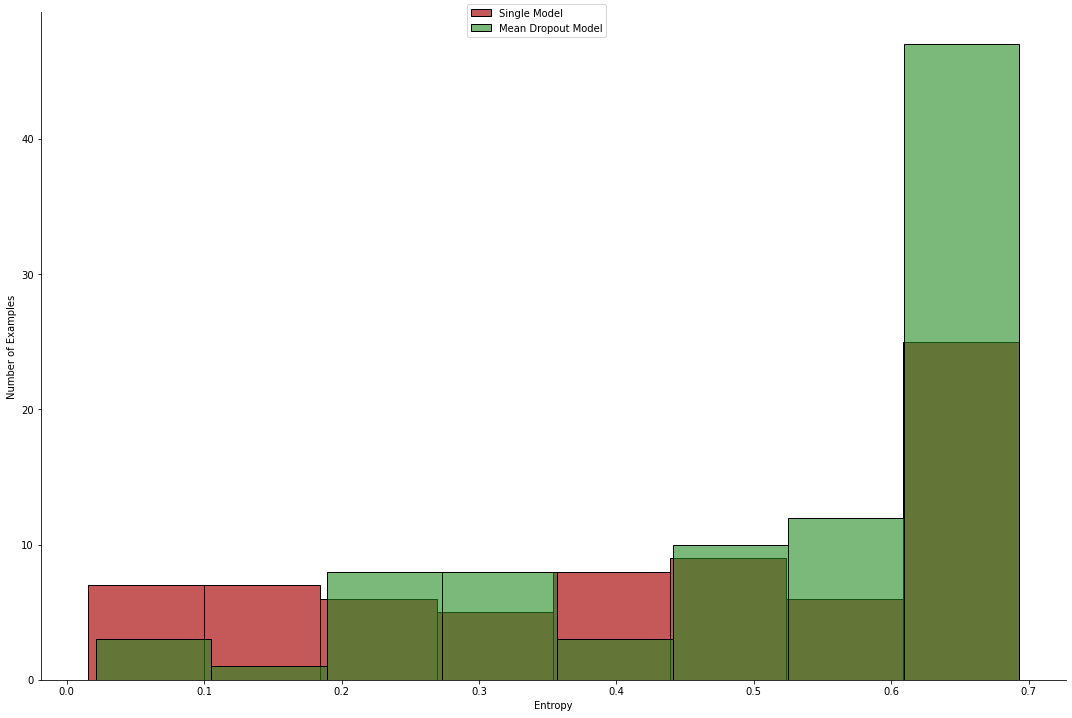}
    \caption{Histogram of entropies for output probabilities of SACDNet and MC-SACDNet.}
    \label{fig:fig5}
\end{figure}

In order to further clarify the reliability of the MC-Dropout extension technique all the examples from the test that were misclassified by either technique were separated and the entropy for the output probabilities was calculated. The histogram of the calculated entropy is given in figure~\ref{fig:fig5}, where each bin represents the number of examples per entropy range. 

It can be observed from the figure~\ref{fig:fig5} that a higher entropy, which counts for a higher uncertainty, is assigned to more misclassified examples by the proposed MC-Dropout-based extension as compared to the simple model. Hence adding MC-Dropout further improves the proposed technique, reducing the number of overconfident wrong inferences.

\section{Conclusion and Future Work}
\label{sec:conclusionfuturework}
Type 2 diabetes mellitus is a chronic disease that is often undiagnosed and can lead to severe complications. It is desirable to have a system that can help in the early diagnosis of T2DM, reducing the yearly health expenses spent tackling complications of undiagnosed T2DM and also saving millions of lives every year that are lost due to late diagnosis of T2DM. This study proposes a novel deep learning-based model to predict T2DM using routine EHR features. SACDNet was able to perform better than the baseline machine and deep learning models. The study also presents a complete real-world solution for building an early diabetes detection system, developing a diverse corpus from real-world EHR data. Also, the proposed framework based on MC-Dropout can help make predictions with more reliability and abstain from predicting in cases where the model is uncertain. The proposed technique and framework can be plugged into existing EHR software with little or no changes. Also, the methodology followed in this study can be used to build predictive software for other diseases like hypertension, ischemic heart disease, etc.

\bibliography{ref}

\end{document}